\documentclass{article} 
\usepackage{nips11submit_e,times}
\usepackage{array}
\usepackage{url}
\usepackage{graphicx}
\usepackage{amsmath}

\DeclareMathOperator*{\argmax}{arg\,max}

\title{Discriminately Decreasing Discriminability with Learned Image Filters}

\author{
Jacob Whitehill and Javier Movellan\\
Machine Perception Laboratory\\
University of California, San Diego\\
La Jolla, CA 92093\\
\texttt{\{ jake, movellan \}@mplab.ucsd.edu} \\
}

\newcommand{\ddtj}{\frac{\partial}{\partial \theta_j}}
\newcommand{\fbar}{\overline{f}}

\newcommand{\xbar}{\overline{x}}
\newcommand{\Xbar}{\overline{X}}

\nipsfinalcopy 

\begin{document}

\maketitle

\begin{abstract}
In machine learning and computer vision, input images are often filtered
to increase data discriminability.
In some situations, however, one may wish to purposely \emph{decrease} discriminability of one classification task
(a ``distractor'' task),
while simultaneously \emph{preserving} information relevant to another (the task-of-interest):
For example, it may be important
to mask the identity of persons contained in face images before submitting them to a crowdsourcing site
(e.g., Mechanical Turk) when labeling them for certain facial attributes. Another example is
inter-dataset generalization: when training on a dataset with a particular covariance structure among
multiple attributes, it may be useful to \emph{suppress} one attribute while preserving another so that a trained
classifier does not learn spurious correlations between attributes. In this paper we present
an algorithm that finds optimal filters to give high discriminability to one task while simultaneously
giving low discriminability to a distractor task.
We present results showing the effectiveness of the proposed technique on both simulated data and natural face images.
\end{abstract}

\section{Introduction}
In machine learning and computer vision, images are commonly filtered prior to classification to enhance
class discriminability. Such filters may consist of manually constructed filters (e.g., low-pass, band-pass filters)
or may be learned directly from the data (e.g., using Deep Belief Networks \cite{LeeEtAl2009} or Independent Components Analysis \cite{BellSejnowski1997}).
However, there also exist scenarios in which it may be
useful to intentionally \emph{decrease} discriminability for \emph{one} classification task (a ``distractor'' task),
while enhancing or at least preserving discriminability for \emph{another} task (the task-of-interest).
Discriminability can pertain to perception by humans, or analysis by a machine classifier. Two scenarios
where such filtering is useful
include (1) preservation of privacy during data labeling, and
(2) generalization to datasets with different correlation structure.

(1) {\em Preservation of privacy}: Machine learning is increasingly making use of crowdsourcing services such as the
Amazon Mechanical Turk, in which not all labelers can be trusted. In some situations, the data to be labeled may
contain sensitive information that should not be released to the public, e.g., the identity of people's faces or
the geographical locations of satellite images. It may be useful to first filter the images before uploading them
to the Mechanical Turk so that identity/location is removed, but so that the task-of-interest remains highly discriminable.
For the case of facial identity removal, this process is known as face de-identification \cite{NewtonEtAl2005}.

(2) {\em Generalization to datasets with different correlation structure}:  
In some training datasets there exist strong correlations between different attributes that can impair generalization performance
to data with different covariance structure (\emph{covariate shift}).
Consider, for example, a classifier, intended to recognize some attribute
A, that is trained on a dataset in which there is a strong correlation between attributes A and B. Such a classifier
may perform very badly when tested on a different dataset in which the correlation between A and B is low or perhaps 
negative. It may be useful, when training a classifier for A, to first filter the training data to \emph{preserve}
discriminability of A, while \emph{suppressing} discriminability of B, so that the spurious correlation between A and B
is not learned.

In this paper, we present a novel algorithm for learning an image filter (parameterized by $\theta$) from labeled data that simultaneously \emph{preserves}
discriminability of the task-of-interest while \emph{suppressing} discriminability of the distractor task.
In this sense, the filter ``discriminately decreases discriminability'' of the images.
In the experiments in the paper, we focus on \emph{image} filters, but in fact the data can be of any dimensional
representation. We focus on discriminating \emph{binary} attributes, but as shown in Section \ref{sec:remove_identity},
suppression of binary gender discrimination also significantly removes face identiability as well.
Before presenting our algorithm in Section \ref{sec:algorithm}, we first provide a simple example of ``discriminately decreasing
discriminability'' in Section \ref{sec:example2d}. The rest of the paper consists of experimental results.


\section{Simple example in $R^2$}
\label{sec:example2d}
\begin{figure}
\begin{center}
\includegraphics[width=5in]{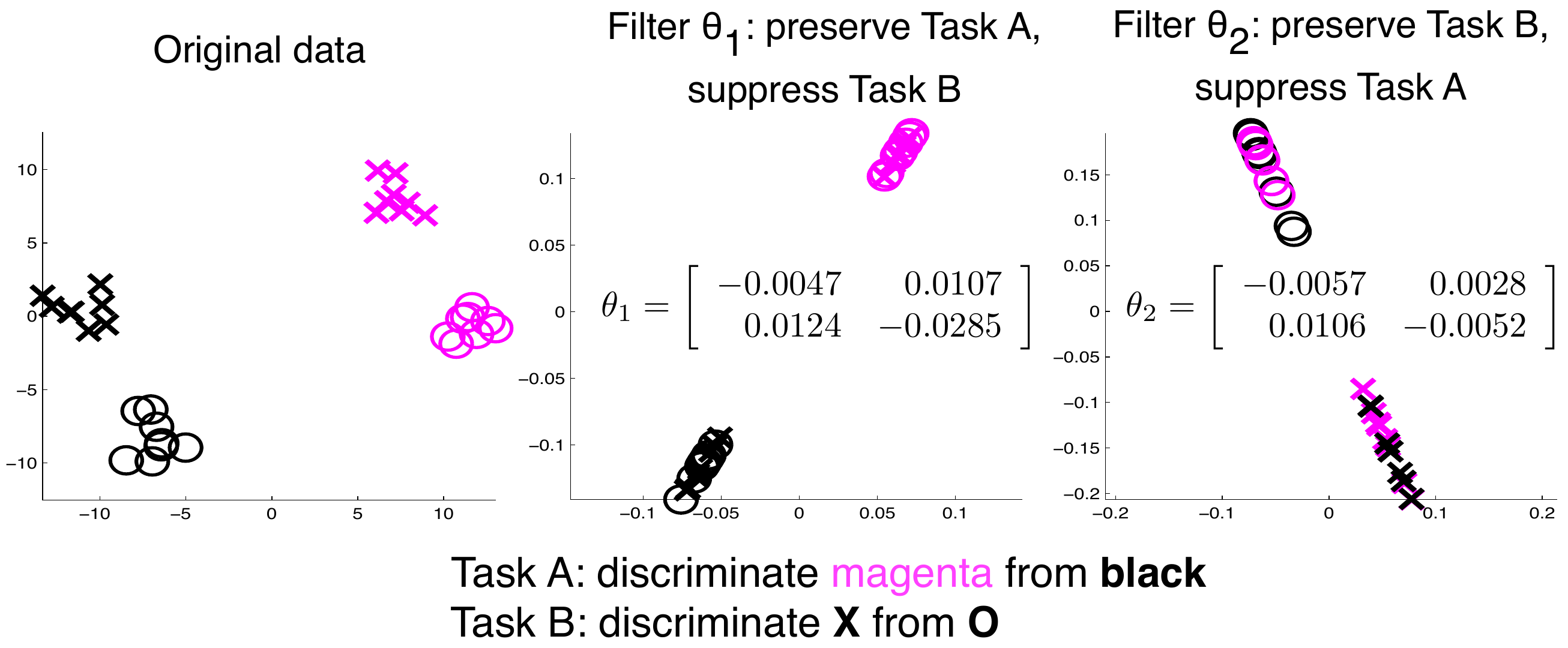}
\end{center}
\caption{A minimal example in $R^2$ showing ({\bf left}) unfiltered data,
data filtered to preserve Task A's and suppress Task B's discriminability ({\bf center}),
and data filtered to suppress Task A's and preserve Task B's discriminability ({\bf right}).
}
\label{fig:example2d}
\end{figure}
Consider the set of 28 data points $\{ x_i \}$ (in $R^2$) shown in Figure \ref{fig:example2d} ({\bf left}):
Each point $x_i$ is given binary labels
for two labeling tasks. Points labeled 0 for Task A are shown in magenta, while points
labeled 1 for task A are black. On the other hand, points labeled 0 for Task B are marked
as crosses, while points labeled 1 are shown as circles. In their unfiltered original form,
both tasks are easily discriminated, as illustrated in Figure \ref{fig:example2d}.

Suppose now that we filter the data using $\theta_1$ (in this case, a general linear transformation), as shown in the {\bf center} part of the figure:
Task A (color) is highly discriminable, while Task B (marker) is not -- the two marker styles (circles
and crosses) appear to overlap.  Similarly, we can use $\theta_2$ to suppress discriminability of
Task A and preserve discriminability of Task B, in which case we arrive at
the filtered points shown in Figure \ref{fig:example2d} ({\bf right}). The goal of the algorithm in this paper is to learn
such linear transformations (filters) automatically.

\section{Algorithm: Learning a filter to discriminately decrease discriminability}
\label{sec:algorithm}
The proposed method requires quantifying data discriminability as $J^*$, as described in the next subsection. The key is that
$J^*$ can be found \emph{analytically} as a function of its input. Using $J^*$, we pose an optimization
problem to maximize the ratio of discriminabilities of Tasks A and B w.r.t.~the filter $\theta$ which
transforms the data.

\subsection{Quantifying discriminability as $J^*$}
The measure of discriminability we use is the ratio of between-class variance to within-class variance, first proposed
by Fisher \cite{Fisher1936} and used in Fisher's Linear Discriminant analysis.

Let $\{ x_i \}$ represent a set of data (column vectors) to be classified, where each $x_i \in \mathcal{R}^d$, and let
$y_i\in \{ 0,1 \}$ represent the binary class label for each $x_i$ for some labeling task.
In our setting, each $x_i$ might be a face image with $d$ pixels, and each $y_i$ might represent, for example,
whether or not the person in image $i$ is smiling.
One useful measure of discriminability of the data w.r.t.~the class labels is Fisher's discriminability
criterion, $J$, which measures the ratio of \emph{between-class variance} $B$ to the \emph{within-class variance} $W$ after projecting the $\{ x_i \}$ onto some direction $p\in R^d$. Depending on the choice
of $p$, the data $\{ x_i \}$ may become more or less discriminable w.r.t.~the class labels $\{ y_i \}$.
When Fisher's linear discriminant is used for actual classification,
the $p$ represents the normal vector to the separating hyperplane of the two classes.

{\bf Notation}: Let $N_0$ denote the number of data vectors $\{ x_i \}$ such that
$y_i=0$, and let $X_0$ denote the $d \times N_0$ matrix formed from the $N_0$
data points in class 0.
We can define $N_1$ and $X_1$ analogously.
We write the mean data vector for class $0$ as $\xbar_0$,
i.e., $\xbar_0 \doteq \frac{1}{N_0}\sum_{i:y_i=0} x_i$ and define $\xbar_1$ analogously.
$\xbar$ is the mean over all $\{ x_i \}$.
Finally, define $\Xbar_0$ (or $\Xbar_1$) as the $d\times N_0$
(or $d\times N_1$) matrix containing $N_0$ (or $N_1$) copies of $\xbar_0$ (or $\xbar_1$).

Given the notation above, Fisher's linear discriminability can be computed as
\begin{equation}
\label{eqn:fisher}
J(X_1,X_0,p) = \frac{p^\top B p}{p^\top W p}
\end{equation}
where the between-class variance is defined as
$B = (\xbar_1 - \xbar_0) (\xbar_1 - \xbar_0)^\top$
and the within-class variance is defined as
$W = (X_1 - \Xbar_1)(X_1 - \Xbar_1)^\top + (X_0 - \Xbar_0)(X_0 - \Xbar_0)^\top$.
$W$ can be regularized as $W_{\textrm{reg}}=\alpha I + (1-\alpha) W$ with regularization parameter $\alpha \in [0,1]$.
For the remainder of the paper we refer to $W_{\textrm{reg}}$ simply as $W$.

One advantage of Fisher's linear discriminant over other classification methods (e.g., 
support vector machines, multivariate logistic regression) is that the optimal $p^*$
that maximizes the discriminability of the $X_1$ from the $X_0$ can be found analytically \cite{Bishop2006}:
\begin{equation}
\label{eqn:best_discriminant}
p^*(X_1,X_0) \doteq \argmax_p J(X_1,X_0,p) = W^{-1} (\xbar_1 - \xbar_0)
\end{equation}
Given this solution for $p^*$, we can define the ``Fisher maximal discriminability'' of $X_1$
and $X_0$ as:
\begin{equation}
\label{eqn:fisher_maximal}
J^*(X_1,X_0) \doteq J(X_1,X_0,p^*(X_1,X_0))
\end{equation}

\subsection{Discriminability for two tasks}
Let us now consider a set of data $\{ x_i \}$ where each $x_i\in\mathcal{R}^d$, as above. However, now we
are interested in \emph{two} sets of class labels for two different binary labeling tasks A and B.
For instance, the $\{ x_i \}$ might represent a set of
face images, and task A might correspond to whether $x_i$ is a smiling face or not, whereas task B might represent whether the face in $x_i$ is male or female. Instead of $X_0$ and $X_1$, we define
$X_{0a}$ and $X_{1a}$ to represent the data points $\{ x_i \}$ that are labeled as class 0 or 1,
respectively, for task A; we define $X_{0b}$ and $X_{1b}$ analogously for task B. Then, the Fisher maximal discriminability
for Task A is $J^*(X_{1a},X_{0a})$ and for Task B is $J^*(X_{1b},X_{0b})$.

\subsection{Finding filter $\theta$ to ensure high $J^*$ for Task A, low $J^*$ for Task B}
Now, suppose that we filter each $x_i$ using any filter
function $\mathcal{F}(\theta, \cdot)$ that is differentiable in $\theta$.
By varying $\theta$, we can change the Fisher maximal discriminability $J^*$ for both tasks.\footnote{
	$J^*$ can be affected by filter $\theta$ even when the linear transformation that the filter induces is invertible.
	In contrast, \emph{linear separability} (existence/non-existence of a separating hyperplane) cannot be
	affected by any invertible linear transformation -- see Supp.~Materials for a proof.
}
Two useful filtering operations include (1)
Convolution: $\mathcal{F}(\theta,x) = x * \theta$, where $\theta$ represents the convolution kernel in vector form; and (2)
pixel-wise ``masking'': $\mathcal{F}(\theta,x) = \textrm{Diag}(\theta) x$ where $\textrm{Diag}(\theta)$ represents
a diagonal matrix formed from the vector $\theta$.
In this case, $\theta$ represents a ``mask''
placed over the image that allows the original image's pixels to pass through with varying strength.
Let us define $f_i(\theta) \doteq \mathcal{F}(\theta,x_i)$ as the output of the filter $\mathcal{F}$ on $x_i$,
and let us define $F_{0a}(\theta),F_{1a}(\theta),F_{0b}(\theta),F_{1b}(\theta)$ analagously 
to their (unfiltered) counterparts $X_{0a},X_{1a},X_{0b},X_{1b}$.

{\bf Goal}: We wish to find the filter parameter vector $\theta$ that gives \emph{high} Fisher maximal
discriminability ($J^*$) to Task A, while simultaneously giving \emph{low} Fisher maximal
discriminability to Task B. This can be formulated as an optimization problem over $\theta$ in several
different ways; we choose the following ``ratio of discriminabilities'' metric $R(\theta)$:
\begin{equation}
R(\theta) \doteq \log \frac{J^*(F_{1b}(\theta), F_{0b}(\theta))}{J^*(F_{1a}(\theta), F_{0a}(\theta))} + \beta\theta^\top \theta
\label{eqn:objective}
\end{equation}
where $\beta\geq 0$ is a scalar regularization parameter on $\theta$.
Since all of the $F(\cdot)$ are differentiable functions of $\theta$, and since $J^*(\cdot,\cdot)$ is given by the simple
formulas in Equations \ref{eqn:fisher_maximal} and \ref{eqn:fisher}, we can use gradient descent to locally minimize
the objective function in Equation \ref{eqn:objective} w.r.t.~$\theta$.
The derivative expressions are given in the Supplementary Materials.



\subsection{Reconstruction from filtered images} 
\label{sec:reconstruction}
The gradient descent procedure described above will find a $\theta$ that locally minimizes
$R(\theta)$, but there is no guarantee that the filtered images $F$ will visually resemble
the original images $X$ or that humans can interpret them. For machine classification (e.g.,
when learning a filter to improve inter-dataset performance), this may not matter, but for human
labeling applications, it may be necessary to ``restore'' the filtered images to a more intuitive form.
Hence, as an optional step, linear ridge regression can be used to convert the filtered images $F$ to
a form more closely resembling the original images $X$, while still 
preserving the property that they are highly discriminability for Task A and not highly discriminable for Task B.
In particular, we can compute the $d\times d+1$ (the extra +1 is for the bias term) linear transformation $P$ that minimizes
\[ \left\| X - P\left[ \begin{array}{c}F\\{\bf 1} \end{array} \right] \right\|_{\textrm{Fr}}^2 + \gamma \left\| P \tilde{I} \right\|_\textrm{Fr}^2 \]
where $\gamma >0$ is a scalar ridge strength parameter,
$\tilde{I}$ is the identity matrix except that the last ($d+1$th) diagonal entry is 0 instead of 1 (so that there is
no regularization on the bias weight), and Fr means Frobenius norm.

The ridge term in the linear reconstruction is critical: because many of the filters that
the gradient descent procedure learns correspond to invertible linear transformations, linear regression
without regularization would transform each $f_i$ back to $x_i$ with no loss of information, which would
defeat the purpose of filtering at all.  With ridge regression, on the other hand, only the
``more discernible'' aspects of the image (i.e., the task-of-interest) are restored clearly,
while the ``less discernible'' aspects (pertaining to the distractor task) are not. By varying $\gamma$,
one can cause each ``reconstructed'' image $g_i$ (where $g_i \doteq P\left[ \begin{array}{c}f_i\\ 1 \end{array} \right]$) to strongly resemble
the mean image $\xbar$ (for large $\gamma$) or to strongly resemble its unfiltered counterpart $x_i$ (for small $\gamma$).
In practice, $\gamma$ is chosen based on visual inspection of the reconstructed training images so that, to the human observer,
the task-of-interest is clearly discriminable while the distractor task is not.

\section{Experiment I: synthetic data}
\label{sec:simulation}
\begin{figure}
\parbox[0]{1.25in}{
	\includegraphics[width=1.25in]{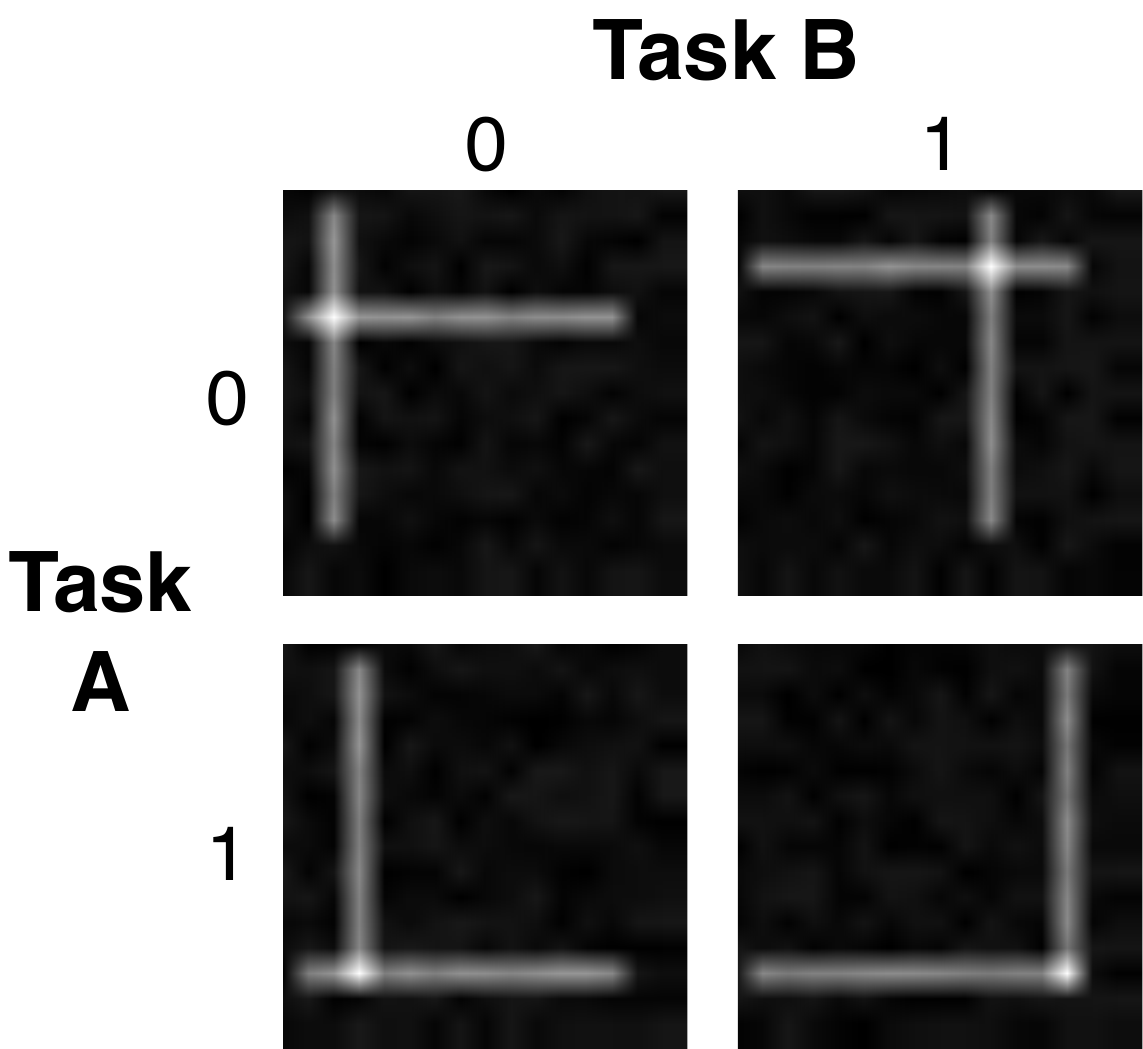}
}
\parbox[1.25]{1.75in}{
	\includegraphics[width=1.75in]{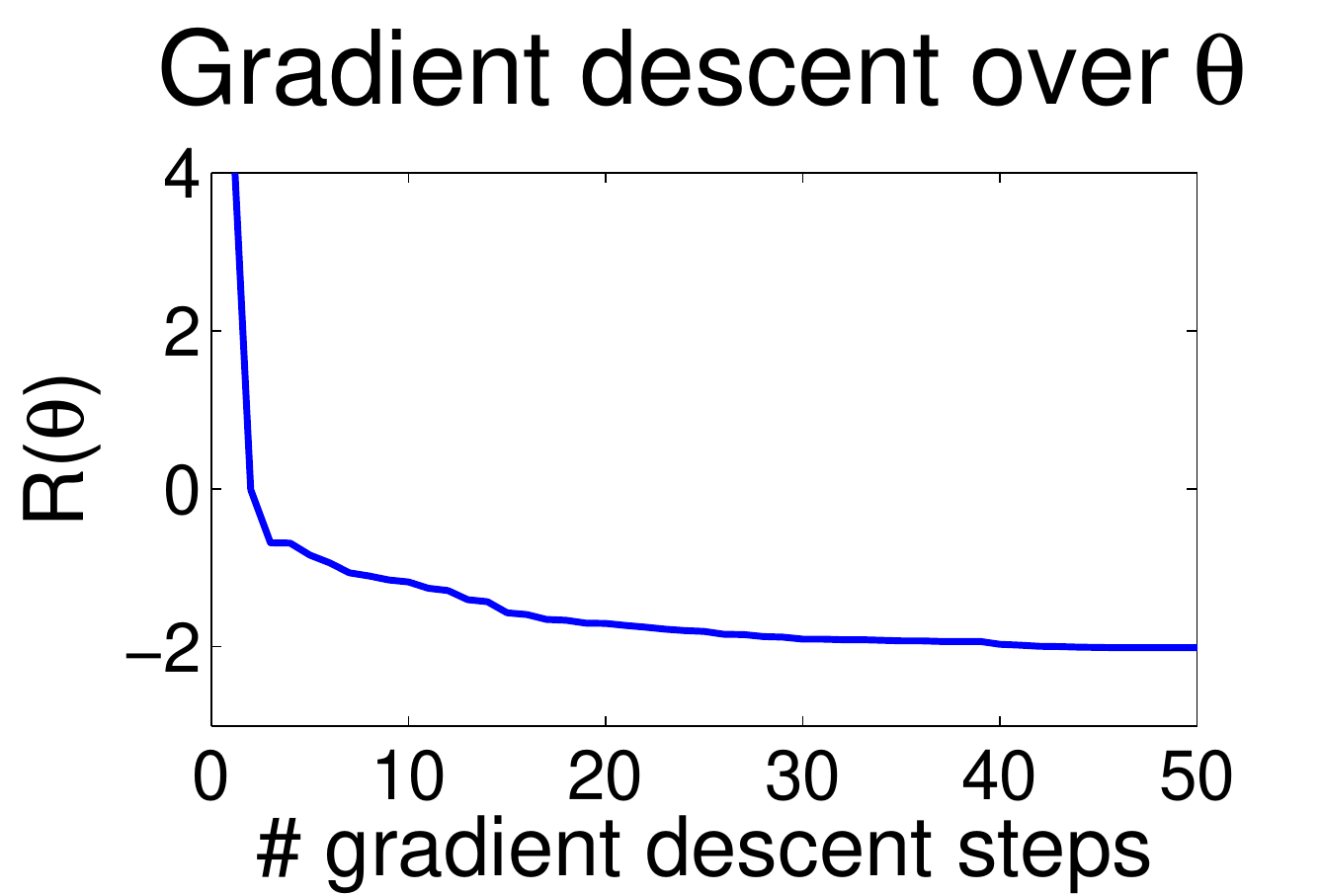} \vspace{0.2cm}
}
\parbox[3.0]{1.5in}{
	\setlength{\tabcolsep}{0.75mm}
	\setlength\fboxsep{0pt}
	\setlength\fboxrule{0.5pt}
	\begin{tabular}{cccccc}
	\multicolumn{6}{c}{Learned convolution kernel ($\theta\in\mathcal{R}^{5\times 5}$)}\\
	\fbox{\includegraphics[width=0.3in]{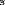}} &
	\fbox{\includegraphics[width=0.3in]{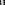}} &
	\fbox{\includegraphics[width=0.3in]{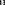}} &
	\fbox{\includegraphics[width=0.3in]{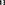}} &
	\fbox{\includegraphics[width=0.3in]{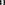}} &
	\fbox{\includegraphics[width=0.3in]{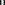}} \\
	0 & 10 & 20 & 30 & 40 & 50 \\
	\multicolumn{6}{c}{\# gradient descent steps} \\
	\end{tabular}
}
\caption{{\bf Left}: Synthetic images consisting of vertical and horizontal lines at different
positions. {\bf Center}: gradient descent curve over $R(\theta)$ to a learn a filter to preserve Task A and
suppress Task B. {\bf Right}: The filters learned at corresponding gradient descent steps.}
\label{fig:example_lines}
\end{figure}

\begin{figure}
\setlength{\tabcolsep}{0.5mm}
\begin{center}
\begin{tabular}{rcccccccc}
Unfiltered patches $\{ x_i \}$ &
\includegraphics[width=0.4in]{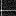} &
\includegraphics[width=0.4in]{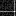} &
\includegraphics[width=0.4in]{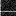} &
\includegraphics[width=0.4in]{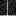} &
\includegraphics[width=0.4in]{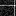} &
\includegraphics[width=0.4in]{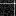} &
\includegraphics[width=0.4in]{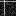} &
\includegraphics[width=0.4in]{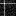} \\
Filtered patches $\{ f_i \}$ &
\includegraphics[width=0.4in]{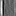} &
\includegraphics[width=0.4in]{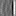} &
\includegraphics[width=0.4in]{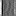} &
\includegraphics[width=0.4in]{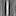} &
\includegraphics[width=0.4in]{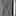} &
\includegraphics[width=0.4in]{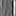} &
\includegraphics[width=0.4in]{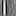} &
\includegraphics[width=0.4in]{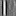}\\
\end{tabular}
\end{center}
\caption{{\bf Top}: unfiltered image patches consisting of superimposed vertical and horizontal
lines plus uniform noise. {\bf Bottom}: the same images filtered with a convolution kernel
designed to suppress discriminability of Task B (horz.~lines) while preserving discriminability
of Task A (vert.~lines).}
\label{fig:example_lines_filtered}
\end{figure}
In our first experiment we studied whether the proposed algorithm could operate on
images ($16\times 16$ pixels) consisting of simple line patterns in order to suppress lines in one direction
while preserving them in another. For the filtering operation, we chose to
learn a convolution kernel of $5\times 5$ pixels. In this study, all images 
contained one horizontal line and one vertical line at random locations:	
In Task A, an image was labeled 0 if it contained a
vertical line in the left half of the image, and it was labeled 1 if its vertical
line was in the right half. In Task B, an image was labeled 0 if its horizontal line was
in the top half, and labeled 1 if it was in the bottom half. Each image $x_i\in R^{16\times 16}$
was generated by adding one vertical and one horizontal line (of pixel intensity 1) at random image positions, and then
adding uniform noise in $U[0,0.5)$ to all pixels in the image.
Example images are shown in Figure \ref{fig:example_lines} ({\bf left}).

After generating 1000 images according to the procedure above, we initialized the 
convolution kernel $\theta\in R^{5\times 5}$ to random values from $U[0,1)$
(shown in Figure \ref{fig:example_lines} as the filter kernel at gradient descent step 0)
and then applied the algorithm above to learn a filter
to preserve Task A while suppressing Task B. We set $\beta$ to 0.5. The descent curve is
show in Figure \ref{fig:example_lines} ({\bf center}), and the learned filter kernel
at every 10 steps is shown below the graph.

After filtering the images using the convolution kernel learned after 50 descent steps, we 
arrived at the images shown in Figure \ref{fig:example_lines_filtered}.
Notice how the horizontal lines have been almost completely
eradicated, thus decreasing class discriminability for Task B.

\section{Experiment II: natural face images}
\label{sec:natural_faces}
\subsection{Preserve expression, suppress gender}
\label{sec:natural_faces_sn}
We applied the proposed filter learning method to natural face images from the GENKI dataset
\cite{GENKI}, which consists of 60,000 images that have been manually labeled for 
2 binary attributes -- smile/non-smile and male/female -- as well as the 2D positions of the
eyes, nose, and mouth, and the 3D head pose (yaw, pitch, and roll). In this experiment
we assessed whether a filter could be learned to \emph{preserve discriminability of expression} (smile/non-smile),
while \emph{suppressing discriminability of gender}. We used a pixel-wise ``mask''
filter (see Section \ref{sec:algorithm}) of the same size as the images ($16\times 16$ pixels).

From the whole GENKI dataset we selected a training set consisting of 1740 images (50\% male and 50\% female; 50\% smile and 50\% non-smile)
whose yaw, pitch, and roll parameters were all within $5\deg$ of frontal. All of the images were registered to a common
face cropping using the center of the eyes and mouth as anchor points. They were then downscaled to a resolution of
$16\times 16$ pixels. In addition, we similarly extracted a separate testing set consisting of 100 images (50 males, 50 females, and 50 smiling, 50 non-smiling)
with the same 3D pose characteristics. The filter $\theta$ was initialized component-wise by sampling from $U[0,1)$.

\begin{figure}
\begin{center}
\includegraphics[width=6in]{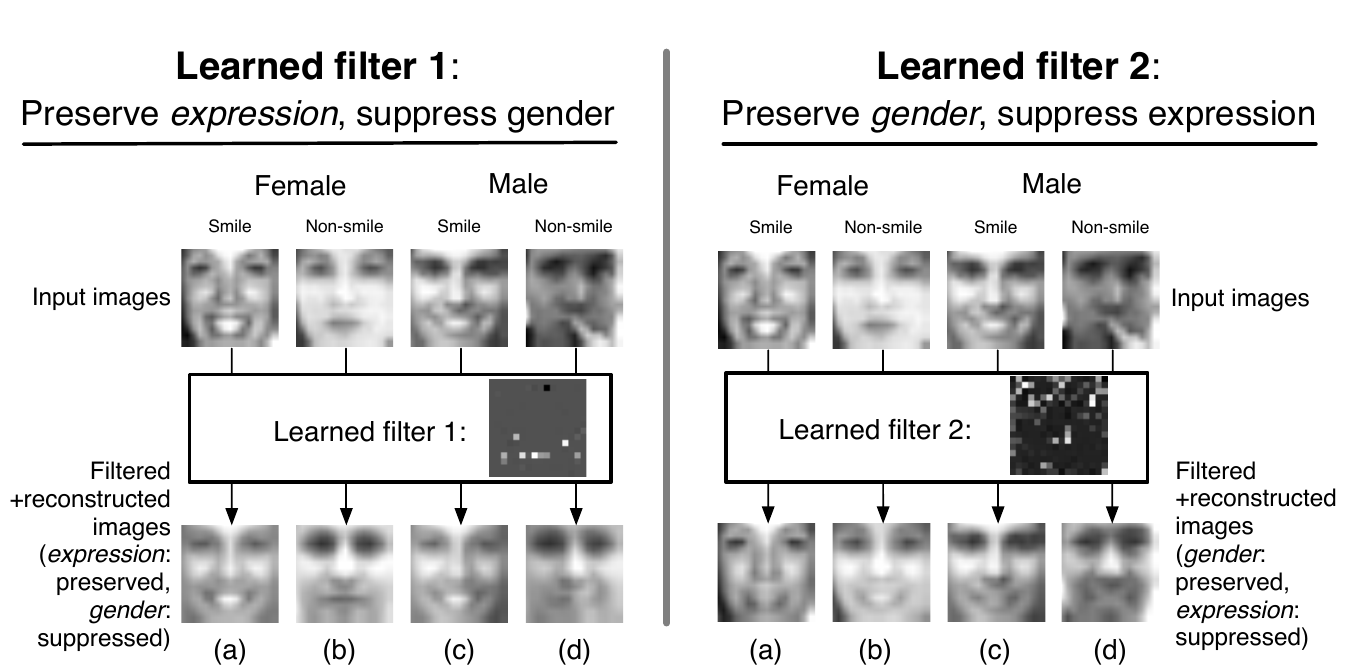}
\end{center}
\caption{Face images from the GENKI dataset that have been filtered to preserve expression and suppress gender ({\bf left}),
or to preserve gender and suppress expression ({\bf right}). Filters were learned using the algorithm presented in
Section \ref{sec:algorithm}. Learned filter masks are shown next to ``Learned filter:''. }
\label{fig:natural_faces}
\end{figure}

Using the training set for learning the filter, and setting the regularization parameter
$\beta=0.5$ ($\alpha=0.1$ as always), we applied conjugate gradient descent for 100 function evaluations. The learned filter was
then applied to all of the training images. Finally, we applied the image reconstruction technique described
in Section \ref{sec:reconstruction} to restore the filtered images to a form more easily analyzable by humans. The reconstruction
ridge parameter $\gamma$ was selected, by looking only at the training images, so that smile appeared well discriminable whereas gender
did not (in this case, $\gamma=6e-2$).
Examples of the input images as well as the filtered (+ reconstructed) images are shown in Figure \ref{fig:natural_faces} ({\bf left}).
The learned filter mask is shown to the right of the text ``Learned filter''.
As shown in the figure, most of the smile information in the filtered images is preserved, and while gender may still be partially discernible, much of the
gender information has been suppressed by the filter.

To assess quantitatively the ability of the learned filter to preserve expression and suppress gender, we posted a labeling task
to the Amazon Mechanical Turk consisting of 50 randomly selected pairs of \emph{filtered} images selected from the \emph{testing} set
using the filter learned according to the above procedure.
Each pair contained 1 smiling image and 1 non-smiling image presented in random order (Left or Right), and the labeler was asked to select which image
-- Left or Right -- was
``smiling more''. The entire set of 50 image pairs was presented to 10 Mechanical Turk workers,
and their opinions on each pair were combined using Majority Vote\footnote{We also applied an algorithm for optimal integration of crowdsourced labels \cite{WhitehillEtAl2009}; 
see Supp.~Materials.}, with ties resolved by selecting the ``Right'' image. Accuracy of
the Mechanical Turk labelers compared to the official GENKI labels was measured
as the probability of correctness on a 2 alternative forced choice task (2AFC), which is equivalent under mild conditions to the
Area under the Receiver Operating Characteristics curve ($A'$ statistic)
that is commonly used in the automatic facial expression recognition literature (e.g., \cite{LuceyEtAl2010}). We similarly generated a set of 50 
randomly selected pairs of filtered images containing 1 male and 1 female. As a baseline, we compared gender and smile labeling accuracy
of the filtered images to similar tasks for the \emph{unfiltered} images. 
Results are shown in Table \ref{tbl:natural_faces_results}.

As shown in the table, the learned image filter substantially reduced discriminability of gender (from $98\%$ to $58\%$),
while maintaining high discriminability of expression ($94\%$ to $96\%$) compared to the baseline (unfiltered) images.

{\bf Comparison to a manually constructed filter}:
In the case of expression and gender attributes, one might reasonably argue that the ``optimal filter'' for preserving
smile/non-smile and suppressing male/female information would be simply to
crop and display only the mouth region of each face.
Hence, we performed an additional experiment in which we compared
Mechanical Turk labeling accuracy on 50 pairs of filtered images, generated similarly as described above,  using a manually
constructed mask filter consisting of just the mouth region (rows 11 through 15 and columns 4 through 13 of each $16\times 16$ face image).
Results are in Table \ref{tbl:natural_faces_results}:
while smile discriminability is equally high as the learned filter 1, gender discriminability using the manually constructed
filter was substantially higher ($74\%$ compared to $58\%$), indicating that the manually constructed filter actually
allowed considerable gender information to pass through. This suggests that
a learned filter can work better than a manually constructed one 
even when strong prior domain knowledge exists.

\begin{table}
\begin{center}
\caption{Accuracy (2AFC) of workers on Mechanical Turk when labeling filtered GENKI images}
\label{tbl:natural_faces_results}
\vspace{0.25cm}
\begin{tabular}{l||r|r}
{\bf Filter method} & {\bf Expression} & {\bf Gender} \\\hline
Unfiltered (baseline) & $94\%$ & $98\%$ \\ \hline
Learned filter 1:
Preserve expression, suppress gender & $96\%$ & $58\%$  \\ \hline
Manually constructed filter:
show mouth region only & $96\%$ & $74\%$ \\ \hline
Learned filter 2:
Preserve gender, suppress expression & $64\%$ & $86\%$ \\
\end{tabular}
\end{center}
\end{table}

\subsection{Preserve gender, suppress expression}
Analogously to Section \ref{sec:natural_faces_sn}, we also learned a filter
to preserve \emph{gender} and suppress \emph{expression}, using an identical training procedure to that described above.
Examples of the filtered (+ reconstructed)
images ($\gamma=9\times 10^{-3}$) are shown in Figure \ref{fig:natural_faces} ({\bf right}). Note how, for face image (b), 
the filter not only ``suppressed'' the expression of the non-smiling female, but actually seems to
``flip'' the smile/non-smile label so that the woman appears to be smiling.
The accuracy compared to baseline (unfiltered)
images is shown in Table \ref{tbl:natural_faces_results}. While accuracy of gender labeling did drop from
$98\%$ to $86\%$, it dropped much more for the smiling labeling ($94\%$ to $64\%$) compared to  unfiltered images.

\section{Experiment III: Preserving privacy in face images (face de-identification)}
\label{sec:remove_identity}
\begin{figure}
\setlength{\tabcolsep}{0.5mm}
\begin{center}
\begin{tabular}{cccccccccc}
\multicolumn{10}{c}{\bf Whose face is this?}\\
\multicolumn{10}{c}{\includegraphics[width=0.5in]{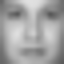}}\\
\multicolumn{10}{c}{Match the filtered face image above to its unfiltered image below.}\\
\includegraphics[width=0.5in]{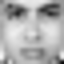} &
\includegraphics[width=0.5in]{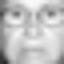} &
\includegraphics[width=0.5in]{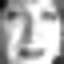} &
\includegraphics[width=0.5in]{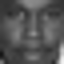} &
\includegraphics[width=0.5in]{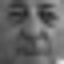} &
\includegraphics[width=0.5in]{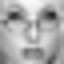} &
\includegraphics[width=0.5in]{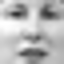} &
\includegraphics[width=0.5in]{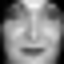} &
\includegraphics[width=0.5in]{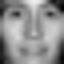} &
\includegraphics[width=0.5in]{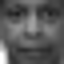}  \\
a & b & c & d & e & f & g & h & i & j\\
\end{tabular}
\end{center}
\caption{{\bf Top}: The preserve-smile, suppress-gender filter both allows smile/non-smile information to pass
through, and also serves as a ``face-identification'' mechanism, as illustrated in the face recognition task above.
The correct face match is (f).}
\label{fig:face_recognition}
\end{figure}

The filters learned in Section \ref{sec:natural_faces} to preserve smile while suppressing gender information
were not designed specifically to suppress the faces' identity. In practice, however, we found that
the identity of the people shown was very difficult to discern in the filtered images. Indeed, it is possible that
gender represents one of the first ``principal components'' of face space, and that, by removing gender, one
implicitly removes substantial identity information as well.

To test the hypothesis that identity was effectively masked by suppressing gender, we created a face recognition
test consisting of 40 questions similar to Figure \ref{fig:face_recognition}: a single face
must be matched to one of 10 unfiltered candidate face images. In half of the
questions, the face to be matched was \emph{filtered} using the preserve-expression, suppress-gender filter
(Section \ref{sec:natural_faces}). In this case, the matching task was very challenging. In the other half of
the questions, the face to be matched was \emph{unfiltered}, and hence the matching task was nearly trivial.
The order of the questions presented to the labelers was randomized, and we obtained results from 10 workers on
the Amazon Mechanical Turk.

{\bf Results}: For the \emph{unfiltered} images, the rate of successful match was $100\%$ for each of the 10 labelers.
For the \emph{filtered} images, the rate of successful match, using Majority Vote, was $15\%$, indicating that
the preserve-smile, suppress-gender filter also removed identity. The
highest successful matching rate of the filtered images for any one labeler was $30\%$. Baseline rate
for guessing was $10\%$.

\section{Experiment IV: Filtering to improve generalization across datasets}
\label{sec:generalization}
Here we provide a proof-of-concept of learning a filter
that improves generalization to novel datasets. Consider a dataset of face images, such as GENKI, with a positive correlation
between gender and smile. If a male/female classifier were trained on these data, then it might learn to distinguish
gender not just by male/female information alone, but also by the correlated presence of smile. When tested on a different
dataset with a different covariance structure, e.g., with negative correlation between smile and gender, the classifier would
likely perform badly. If we first filter the data to suppress smile information but preserve gender information, then the
trained classifier might not suffer when applied to the new dataset.

To test this hypothesis, we partitioned the GENKI images used in Section \ref{sec:natural_faces} into a training set (4062 images)
and a testing set (970 images). As before, all images were $16\times 16$ pixels.
In the training set, the correlation between smile and gender was $+0.64$, whereas in the testing set,
it was $-1$. We then trained two support vector machine classifiers with radial basis function (RBF) kernels to classify
gender. One classifier was trained on \emph{filtered} training images, using the gender-preservation, smile-suppression filter
learned in Section \ref{sec:natural_faces}, and the other was trained on \emph{unfiltered} images. The RBF width $\gamma$
was optimized independently ($\gamma \in \{ 10^{-8}, 10^{-7}, \ldots, 10^{+4} \}$)
for each classifier using a ``holdout'' set (a randomly selected $20\%$ subset of the training images). The classifier
trained on unfiltered images was then applied to the unfiltered testing set, and the classifier trained on filtered images
was applied to the filtered testing set.

{\bf Results}: Filtering the images using the gender-preservation, smile-suppression filter
resulted in substantially increased generalization performance: 2AFC accuracy was $0.92$ for the SVM trained on filtered images,
whereas it was only $0.79$ for the SVM trained on unfiltered images.

\section{Related work}
We are unaware of any work that specifically learns filters to simultaneously preserve and suppress different image attributes.
However, the approach taken in this paper is somewhat reminiscent of work by Birdwell and Horn \cite{BirdwellHorn1990},
in which an optimal combination of a fixed set of filters is learned to minimize the conditional entropy
of class labels given the filtered inputs.

In terms of applications to data privacy, our method is related to ``face de-identification'' methods such as
\cite{NewtonEtAl2005,GrossEtAl2006,GrossEtAl2005}. Such methods identify faces which are similar either in terms
of pixel space (\cite{NewtonEtAl2005,GrossEtAl2005}), eigenface space (\cite{NewtonEtAl2005}), or Active
Appearance Model parameters (\cite{GrossEtAl2006}), and then replace clusters of $k$ similar faces with their mean
face, thus guaranteeing that no face can be identified more specifically than to a cluster of $k$ candidates. However,
in contrast to our proposed algorithm,  these methods cannot be ``reversed'' to maximally preserve identity while
minimizing discriminability of a given face attribute.

For the application of generalizing to datasets with different image statistics,
our work is related to the problem of covariate shift \cite{Shimodaira2000} and
the field of transfer learning \cite{PanYang2010}.
The method proposed in our paper is useful when dataset differences are known a priori -- the learned filter
helps to overcome covariate shift by altering the underlying images themselves.

\section{Summary}
\label{sec:conclusions}
We have presented a novel method for learning filters that can preserve binary discriminability for the task-of-interest, while
suppressing discriminability for a distractor task. The effectiveness of the approach was demonstrated on synthetic as well
as natural face images. Interestingly, the suppression of gender implicitly removed considerable facial identity information,
which renders the technique useful for labeling tasks where personal identity should remain private. Finally, we demonstrated
that ``discriminately decreasing discriminality'' may help classifiers to generalize across datasets.

\section*{Supplementary Materials}
\subsection{Proof: Class Separability Unaffected by Invertible Linear Transformation}
Let sets $\mathcal{X}_1$ and $\mathcal{X}_0$ contain the data points (column vectors) for classes 1 and 0, respectively.
Let $T$ be an arbitrary invertible linear transformation. We define \emph{separability} of $\mathcal{X}_1$ and
$\mathcal{X}_0$ to mean that
there exists a hyperplane, with normal vector $w$, such that $w^\top x_i > w^\top x_j$ for every $x_i\in\mathcal{X}_0$
and every $x_j\in\mathcal{X}_1$. {\em Claim}: 
\[ \exists w:\ w^\top x_i > w^\top x_j \quad \iff \quad \exists \widehat{w}:\ \widehat{w}^\top (Tx_i) > \widehat{w}^\top (Tx_j)
\]

\noindent Proof of $\Rightarrow$:
There exists $w$ such that $w^\top x_i > w^\top x_j$ for every $x_i\in\mathcal{X}_0$ and every $x_j\in\mathcal{X}_1$.
Since $T$ is invertible, the matrix $T^{-1}$ exists.
Hence, there exists $w$ such that $w^\top T^{-1} T x_i > w^\top T^{-1} T x_j$. We can define $\widehat{w}^\top \doteq w^\top
T^{-1}$. Then,
$\widehat{w}^\top (T x_i) > \widehat{w}^\top (T x_j)$  for every $x_i\in\mathcal{X}_0$ and every $x_j\in\mathcal{X}_1$.
\\

\noindent Proof of $\Leftarrow$:
There exists $\widehat{w}$ such that $\widehat{w}^\top (Tx_i) > \widehat{w}^\top (Tx_j)$ for every $x_i\in\mathcal{X}_0$ and every $x_j\in\mathcal{X}_1$. Then we can define $w^\top\doteq \widehat{w}^\top T$, and we have
$w^\top x_i > w^\top x_j$ for every $x_i\in\mathcal{X}_0$ and every $x_j\in\mathcal{X}_1$.

\subsection{Experiment II: natural face images -- supplementary results}
For the natural face image labeling tasks, in addition to combining opinions of the 10 workers on Mechanical Turk
using Majority Vote, we also tried using a recently developed method by Whitehill, et.~al \cite{WhitehillEtAl2009}
for combining multiple opinions when the quality of the labelers is unknown a priori. Their algorithm is called
GLAD (Generative model of Labels, Abilities, and Difficulties) and its source code is available online.

In general the results were similar to Majority Vote:

\begin{tabular}{l||r|r}\hline
\multicolumn{3}{c}{\bf Accuracy (2AFC) of labels from Amazon Mechanical Turk}\\
\multicolumn{3}{c}{\bf using GLAD \cite{WhitehillEtAl2009} to combine opinions}\\\hline\hline
{\bf Filter method} & {\bf Expression} & {\bf Gender} \\\hline
Unfiltered (baseline) & $94\%$ & $98\%$ \\ \hline
Learned filter 1: & & \\
Preserve expression, suppress gender & $94\%$ & $54\%$  \\ \hline
Manually constructed filter: & & \\
(show mouth region only) & $98\%$ & $72\%$ \\ \hline
Learned filter 2: & & \\
Preserve gender, suppress expression & $64\%$ & $92\%$ \\
\end{tabular}

\subsection{Derivatives expressions for gradient descent}
Let $\theta_j$ represent the $j$th component of vector $\theta$. 
We abbreviate $p^*(F_1,F_0)$ as $p^*$. Let us also abbreviate each matrix $F(\theta)$ as simply $F$.
Finally, let $\fbar_0(\theta)$ and $\fbar_1(\theta)$ represent the mean \emph{filtered} vectors (with filter $\theta$) for 
class 0 and 1, corresponding to $\xbar_0$ and $\xbar_1$, respectively.

To compute $\frac{\partial R}{\partial \theta_j}$, we can apply the chain rule several times in succession. Most
of the derivatives are relatively straightforward to derive using standard formulas from linear algebra; however,
we present derivatives for the most important terms:
\begin{eqnarray*}
\ddtj R(\theta) &=& \frac{\ddtj J^*(F_{1a},F_{0a})}{J^*(F_{1a},F_{0a})} - \frac{\ddtj J^*(F_{1b},F_{0b})}{J^*(F_{1b},F_{0b})}\\
                &=& \frac{\ddtj J(F_{1a},F_{0a},p^*(F_{1a},F_{0a}))}{J(F_{1a},F_{0a},p^*(F_{1a},F_{0a}))} - \frac{\ddtj J(F_{1b},F_{0b},p^*(F_{1b},F_{0b}))}{J(F_{1b},F_{0b},p^*(F_{1b},F_{0b}))}\\
\ddtj J(F_0,F_1,{p^*}) &=& \frac{\ddtj \left( {p^*}^\top B {p^*} \right)}{{p^*}^\top W {p^*}} -
                       \frac{{p^*}^\top B {p^*}}{({p^*}^\top W {p^*})^2} \ddtj \left({p^*}^\top W {p^*}\right)\\
\ddtj p^* &=& \ddtj \left( B^{-1} (\fbar_1(\theta) - \fbar_0(\theta)) \right)\\
          &=& -B^{-1} \left( \ddtj B \right) B^{-1} (\fbar_1(\theta) - \fbar_0(\theta)) + B^{-1} \left(\ddtj \fbar_1(\theta) - \ddtj \fbar_0(\theta)\right)\\
\ddtj \fbar_k(\theta) &=& \ddtj \left( \frac{1}{N_k} \sum_{i: y_i=k} f_i(\theta) \right)\\
                      &=& \frac{1}{N_k} \sum_{i: y_i=k} \ddtj f_i(\theta) \\
\end{eqnarray*}
The derivatives for $\ddtj f_i(\theta)$ depend on the particular kind of filter. In the subsections below
we find the derivatives for a convolution filter, and a pixel-wise ``mask'' filter.

\subsection*{Derivatives of linear convolution filter}
For the case of convolving two 1-D functions $f$ and $g$ whose domains are both $\mathcal{R}$ (all real numbers),
differentiating the convolution operator is trivial: $\frac{d}{dg} (f*g)=f$. However, in our case we are
interested in finite, discrete convolution of a convolution kernel and an image. Consider for the moment
the case of 1-D convolution of a 3-element kernel $\theta$ with a 3-element image $x$:
\begin{eqnarray}
\theta &=& \left[ \begin{array}{ccc} a & b & c \end{array} \right]\\
x      &=& \left[ \begin{array}{ccc} r & s & t \end{array} \right]\\
\theta*x&=& \left[ \begin{array}{ccccc} ar & as + br & at + bs + cr & bt + cs & ct \end{array} \right]\\
\end{eqnarray}
For the purposes of gradient descent, it is necessary to ``clip'' the convolution operation's output so that
it retains the same size as the input $x$; hence, we define:
\begin{eqnarray}
\textrm{Clip}(\theta*x) &=& \left[ \begin{array}{ccc} as + br & at + bs + cr & bt + cs \end{array} \right]\\
\end{eqnarray}

We can now differentiate $\textrm{Clip}(\theta*x)$ w.r.t.~each dimension of $\theta$:

\begin{eqnarray}
\frac{\partial}{\partial a} (\textrm{Clip}(\theta*x)) &=& \left[ \begin{array}{ccc} s & t & 0 \end{array} \right]\\
\frac{\partial}{\partial b} (\textrm{Clip}(\theta*x)) &=& \left[ \begin{array}{ccc} r & s & t \end{array} \right]\\
\frac{\partial}{\partial c} (\textrm{Clip}(\theta*x)) &=& \left[ \begin{array}{ccc} 0 & r & s \end{array} \right]\\
\end{eqnarray}
Hence, the derivatives of the clipped convolution are computed by ``sliding'' the row vector $x$ across a row of 0's and
clipping at the appropriate indices. For the case of finite discrete 2-D convolution, the situation is analogous -- the
gradient of the 2-D convolution $\theta * x$ is found by ``sliding'' the image matrix $x$ over a matrix of 0's.

While it is possible to specify in mathematical notation the exact indices used for sliding and clipping, it is tedious and
unilluminating. Instead we provide a snippet of Matlab code for computing $\frac{\partial}{\partial \theta}\textrm{Clip}(\theta*x)$ for
2-D image convolution with 0-padding (no wrap-around):
\footnotesize
\begin{verbatim}
function dConvdtheta = computedConvdtheta (x, theta, i)
% dConvdtheta = COMPUTEDCONVDTHETA (x, theta, i)
% computes the derivative with respect to theta_i
% of conv(x, theta).

     m = sqrt(length(theta));            % theta corresponds to m x m convolution kernel
     n = sqrt(length(x));                % x corresponds to n x n image
     idxs = reshape(1:length(theta), [ m m ]);
     [r,c] = find(idxs == i);
     
     xim = reshape(x, [ n n ]);
     dConvdthetaim = zeros(m+n-1, m+n-1);
     dConvdthetaim(r:r+n-1, c:c+n-1) = xim;
     
     % Trim the matrix back down to only be the "center" part of the convolution result
     upperPadding = ceil((size(dConvdthetaim, 1) - n) / 2);
     leftPadding = ceil((size(dConvdthetaim, 2) - n) / 2);
     lowerPadding = size(dConvdthetaim, 1) - n - upperPadding;
     rightPadding = size(dConvdthetaim, 2) - n - leftPadding;
     dConvdthetaim = dConvdthetaim(1+upperPadding:end-lowerPadding,
                                   1+leftPadding:end-rightPadding);
     dConvdtheta = dConvdthetaim(:);
end

\end{verbatim}
\normalsize

\subsection*{Derivatives of element-wise ``mask'' filter}
If we let the filter $\mathcal{F}(\theta,x)=D(\theta)x$, where $D(\theta)$ is a diagonal matrix formed from vector 
$\theta$, then
\begin{eqnarray*}
\ddtj f_i(\theta) &=& \ddtj \mathcal{F}(\theta,x_i)\\
                  &=& \ddtj\left( D(\theta) x_i \right)\\
                  &=& I_j x_i
\end{eqnarray*}
where $I_j$ consists of all 0's except the $(j,j)$th entry which is 1.

{\small
\bibliographystyle{ieee}
\bibliography{paper}
}

\end{document}